\documentclass{article}

\usepackage{PRIMEarxiv}

\usepackage[utf8]{inputenc} % allow utf-8 input
\usepackage[T1]{fontenc}    % use 8-bit T1 fonts
\usepackage{hyperref}       % hyperlinks
\usepackage{url}            % simple URL typesetting
\usepackage{booktabs}       % professional-quality tables
\usepackage{amsfonts}       % blackboard math symbols
\usepackage{nicefrac}       % compact symbols for 1/2, etc.
\usepackage{microtype}      % microtypography
\usepackage{lipsum}
\usepackage{fancyhdr}       % header
\usepackage{graphicx}       % graphics
\graphicspath{{media/}}     % organize your images and other figures under media/ folder
\usepackage{algorithm}
\usepackage{algpseudocode}
\usepackage{amsmath}
\usepackage{amssymb}
\usepackage{bm,times}
\usepackage{subcaption}
\usepackage{multirow}

\usepackage{array}
\usepackage{float}

\newcolumntype{?}{!{\vrule width 1.5pt}}
\graphicspath{{figures/}}

\title{Towards Fair Face Verification: An In-depth Analysis of Demographic Biases}

\author{
  Ioannis Sarridis \\
  Centre for Research and Technology Hellas \\
  Thessaloniki, Greece\\
  \texttt{gsarridis@iti.gr} \\
  \And
   Christos Koutlis \\
   Centre for Research and Technology Hellas \\
   Thessaloniki, Greece \\
   \texttt{ckoultis@iti.gr} \\
   \And
   Symeon Papadopoulos \\
   Centre for Research and Technology Hellas \\
   Thessaloniki, Greece \\
   \texttt{papadop@iti.gr} \\
   \And
   Christos Diou \\
   ~~~~~~~~~Harokopio University of Athens~~~~~~~~~\\
   Athens, Greece \\
   \texttt{cdiou@hua.gr} \\
}

\begin{document}
\maketitle

%%%%%%%%% BODY TEXT
\begin{abstract}
Deep learning-based person identification and verification systems have remarkably improved in terms of accuracy in recent years; however, such systems, including widely popular cloud-based solutions, have been found to exhibit significant biases related to race, age, and gender, a problem that requires in-depth exploration and solutions. This paper presents an in-depth analysis, with a particular emphasis on the intersectionality of these demographic factors. Intersectional bias refers to the performance discrepancies w.r.t. the different combinations of race, age, and gender groups, an area relatively unexplored in current literature. Furthermore, the reliance of most state-of-the-art approaches on accuracy as the principal evaluation metric often masks significant demographic disparities in performance. To counter this crucial limitation, we incorporate five additional metrics in our quantitative analysis, including disparate impact and mistreatment metrics, which are typically ignored by the relevant fairness-aware approaches. Results on the Racial Faces in-the-Wild (RFW) benchmark indicate pervasive biases in face recognition systems, extending beyond race, with different demographic factors yielding significantly disparate outcomes. In particular, Africans demonstrate an 11.25\% lower True Positive Rate (TPR) compared to Caucasians, while only a 3.51\% accuracy drop is observed. Even more concerning, the intersections of multiple protected groups, such as African females over 60 years old, demonstrate a +39.89\% disparate mistreatment rate compared to the highest Caucasians rate. By shedding light on these biases and their implications, this paper aims to stimulate further research towards developing fairer, more equitable face recognition and verification systems.

\end{abstract}
\section{Introduction}
% define the task
Facial recognition technology has transformed numerous sectors, from banking to communication, by providing effective mechanisms for identifying individuals \cite{parkhi2015deep}. One of the key tasks in this domain is face verification, a metric learning problem for determining whether two images depict the same person \cite{schroff2015facenet}. Following the digitalization of society, face verification systems have seen widespread adoption and integration into various applications, bolstered by breakthroughs in computer vision and machine learning \cite{liu2016large,deng2019arcface}.
% the problem of bias in fv
Despite the rapid advancements in such technologies, substantial issues persist, with bias and discrimination being one of them. Bias in face verification systems refers to the performance discrepancies that emerge among diverse demographic groups \cite{buolamwini2018gender,fabbrizzi2022survey,ntoutsi2020bias}. These disparities pose serious ethical concerns and practical limitations on the utility and fairness of face verification systems.

% the literature progress on this
Recent research has largely focused on studying and mitigating racial bias in face verification systems. Pioneering works in this area have shed light on the concerning discrepancies in verification performance between different racial groups \cite{wang2020mitigating,liu2019fair,wang2021meta}. While race is undoubtedly an important factor, the narrow focus on racial bias leaves out potentially significant biases along other demographic dimensions such as gender and age, and their intersections. This oversight creates a limited understanding of bias in face verification systems.
Furthermore, the prevalent measure used to evaluate performance in these studies is accuracy, a general metric that cannot reflect the model's fairness-specific behavior in a comprehensive manner. While measuring the accuracy across different groups is a useful indicator of overall system performance, it has limitations in capturing the specific discrepancies experienced by minority groups. In other words, high accuracy per protected group may conceal significant biases in system performance, rendering them invisible to researchers and practitioners. For instance, it might obscure intersectional biases, where individuals belonging to multiple minority groups experience greater inaccuracies, while also hiding disparities in False Positive Rates (FPR) and False Negative Rates (FNR), which could disproportionately affect certain demographic groups.

%what we did
In this paper, we extend the analysis of bias in face verification systems beyond the dimension of race, to include gender and age-related biases, while also pushing the boundaries of existing research by exploring the interaction of demographic factors, i.e. intersectional bias, by studying race-gender and race-age intersections. To this end, we further annotated the Racial Faces in the Wild (RFW) dataset w.r.t. to gender and age, as the original RFW involves only racial labels. Then we conducted an extensive analysis for all different races, genders, ages and their possible intersections. We also computed several metrics, such as accuracy, mean Average Precision (mAP), True Positive Rates (TPR), FPR, Differences in False Positive Rates (DFPR),  and Differences in False Negative Rates (DFNR), t-distributed Stochastic Neighbor Embedding (t-SNE) \cite{van2008visualizing} visualizations, and  inter/intra-group similarity statistics across demographics. Additionally, we investigated an often-overlooked problem: the significant performance discrepancies in verifying image pairs that depict the same individual but at vastly different ages. For instance, an identification photo taken at the age of 20 compared with a selfie photo captured at the age of 40. This is a frequent occurrence in real-world applications but has scarcely been addressed in the current literature. 
% the results of the analysis
Our analysis reveals a significant level of bias in face verification systems across many of the studied demographic factors and their intersections, which underlines the importance of considering multiple protected attributes simultaneously when assessing or mitigating bias in such systems. 
Finally, we uncover substantial performance differences for image pairs representing the same individual at different ages. This provides valuable insights for the research community to drive the development of face verification methodologies that are more robust and accurate across diverse age ranges, thereby enhancing their fairness and applicability.
\\
The main contributions of this paper are the following:
\begin{itemize}
\item Broadening the scope of bias assessment in face verification systems to include race, gender, age, and their intersections. This contributes to a more comprehensive understanding of the biases inherent in these systems.
\item  Employing mAP, TPR, FPR, disparate impact, and disparate mistreatment metrics to assess bias. By doing so, we reveal larger performance discrepancies than those typically reported by accuracy-based evaluations.
\item Providing t-SNE visualizations and inter/intra-group similarity statistics across demographics. These visual aids and statistical analyses on feature space allow us to explore and identify intricate patterns and potential biases that might not be evident otherwise. For instance, a group presenting high inter-group similarity suggests the model might struggle to differentiate between individuals belonging to this group, underscoring the necessity for applying a bias mitigation approach that specifically focuses on this issue.
\item Enriching the Racial Faces in the Wild (RFW) benchmark with annotations for gender and age. This enhancement aims to inspire and enable researchers to develop methodologies  that consider and address bias w.r.t. multiple protected attributes.
\footnote{The annotations are available at \url{https://www.github.com/gsarridis/rfw-gender-age}.}
% \footnote{The annotations will be publicly available on GitHub upon acceptance.}.

\item Highlighting and quantifying the  performance disparities when verifying image pairs of the same individual taken at significantly different ages – a frequently occurring yet overlooked scenario in real-world applications.
\end{itemize}

% The remainder of this paper is organized as follows: Section 2 provides an overview of related work, Section 3 presents our methodology, and the results of the conducted analysis are detailed in Section 4. Finally, we conclude the paper in Section 5.

\section{Related Work}

% \subsection{Bias in Face Verification}

% Numerous recent studies highlight the significant impact of demographic factors on the performance of face recognition systems \cite{wang2021deep,garcia2019harms,buolamwini2018gender}. The authors of \cite{leslie2020understanding} discuss the ethical issues surrounding the use of face recognition technologies, focusing on the biases and injustices that can arise from their use.

Several recent works have shed light on the profound influence of demographic characteristics on face recognition system performance \cite{buolamwini2018gender,garcia2019harms,leslie2020understanding,wang2021deep}. %Leslie et al. \cite{leslie2020understanding} discuss the ethical concerns linked to face recognition technologies, emphasizing the potential biases and injustices resulting from their application.
Studies focusing on racial bias highlight that faces belonging to certain races tend to be associated with weaker performance even when they are equally represented in training data \cite{wang2019racial,krishnapriya2020issues}. Investigations on the gender attribute show a diminished recognition accuracy for female faces compared to males, even in scenarios where data distributions are biased in favor of female faces \cite{albiero2020face,albiero2020does}.  Regarding age, studies have found that systems perform worse when identifying children's faces \cite{deb2018longitudinal,michalski2018impact}.
Despite these insights, fairness exploration of intersecting demographic attributes remains noticeably limited. In particular, an analysis of intersectional bias in facial analysis models (e.g., smiling, wearing makeup, etc.) by generating manipulated synthetic samples w.r.t. gender, hair length, age, and facial hair is provided in \cite{balakrishnan2021towards}. The face acquisition systems' performance w.r.t. combined demographic attributes is briefly discussed in \cite{cook2019demographic}. Gender-age accuracy disparities in facial attribute extraction and face recognition systems are analyzed in \cite{georgopoulos2020investigating}. Demographic imbalance w.r.t. gender and races in face recognition benchmarks is highlighted in \cite{hupont2019demogpairs}. Finally, Srinivas et al. \cite{srinivas2019exploring} investigated performance discrepancies in face recognition w.r.t. gender, specifically for children.
However, to our knowledge, there has not yet been a systematic analysis of intersectional bias pertaining to gender, age, and race across multiple fairness metrics in face recognition or verification systems.

Regarding bias mitigation, Robinson et al. \cite{robinson2020face} show that using a global threshold for all subgroups in face recognition systems can lead to significant performance differences, and propose the use of subgroup-specific thresholds to balance performance across different demographics. %, thereby mitigating the bias. 
Similarly, a threshold-based bias mitigation approach is introduced in \cite{terhorst2020comparison}.
Solutions for racial bias by learning disparate margins in feature space for different races are proposed in \cite{wang2020mitigating,wang2021meta,yang2021ramface}.
In \cite{dhar2021pass}, a method is introduced that discourages the model from encoding racial information, while
\cite{dhar2020towards} suggests an approach for learning gender-invariant representations for face verification.  Finally, a mutual information-based approach for disassociating representations from a certain protected attribute is introduced in \cite{sarridis2023flac}.
While these works have significantly enriched our understanding of bias in face recognition and provide valuable mitigation strategies, there remains a critical requirement to discover, assess, and mitigate potential bias found in combined protected attributes.

% \subsection{Cross-age Face Verification}
Furthermore, Cross-Age Face Recognition (CAFR) is a very challenging task that often occurs in practical face recognition and verification settings. Techniques such as synthesizing images with the targeted age using generative probabilistic models \cite{nhan2017temporal} and identity-preserved conditional generative adversarial networks \cite{wang2018face} have been explored to address CAFR. Alternatively, some works have focused on separating age and identity components to extract age-invariant representations \cite{wen2016latent,zheng2017age}. However, CAFR remains an under-investigated domain mostly due to the scarcity of proper datasets and its inherent complexity. In this paper, we highlight this issue as part of the highly popular Racial Faces in the Wild (RFW) benchmark dataset \cite{wang2019racial}.

\section{Methodology}
\subsection{Face Recognition vs Verification}
\label{sec:problem_form}
Face recognition can be framed as a classification problem, where the aim is to assign an input image to one of $C$ classes. If we represent an input face image as $\mathbf{X}_i$ and the corresponding label as $y_i$, where $i = 1, \ldots,N$, the objective of a face recognition model $f(\cdot)$ can be written as $f(\mathbf{X}_i) = y_i$, for $i = 1,\dots,N$.
In contrast, face verification is a binary classification task, which involves determining whether two input images depict the same individual. If we represent two input images as $\mathbf{X}_i$ and $\mathbf{X}_j$ and their labels as $y_i$ and $y_j$ respectively, the objective of a face verification model $g(\cdot,\cdot)$ can be written as:
\begin{equation}
g(\mathbf{X}_i, \mathbf{X}_j) =
\begin{cases}
    1, & \text{if}\ y_i = y_j \\
    0, & \text{otherwise}
\end{cases}
\end{equation}
Typically, the feature embeddings extracted by a face recognition model are used as input to $g(\cdot,\cdot)$, which applies a distance measurement between these two representations: 
\begin{equation}
    d(\mathbf{d}_i, \mathbf{d}_j) = \sqrt{\sum_{k=1}^{n} (d_{i,k} - d_{j,k})^2},
\end{equation}
where $d(\mathbf{d}_i, \mathbf{d_j})$ denotes the euclidean distance between the feature vectors $\mathbf{d}_i$ and $\mathbf{d}_j$. Note that also other metrics can be employed, e.g., cosine similarity. Then, the prediction is made by comparing the computed distance with a given threshold.
\subsection{Training Datasets}

Among existing datasets for face recognition and verification, BalancedFace-4 can be considered as the highest quality  in terms of fairness \cite{wang2021meta}. BalancedFace-4 contains approximately 1.3 million images from 28,000 celebrities, with the images representing an approximate balance with respect to skin tones. It was collected by downloading face images from diverse regions worldwide according to a one-million FreeBase celebrity list. The images were meticulously cleaned both automatically and manually by the authors \cite{wang2021meta}, in a manner akin to other prominent face recognition datasets such as VGGFace2 \cite{cao2018vggface2} and MegaFace \cite{kemelmacher2016megaface}.
The images are divided into four distinct bins, labeled as ``Tone I-IV'', where the lightest skin tone is represented as Tone-I and the darkest as Tone-IV based on the Individual Typology Angle (ITA) \cite{chardon1991skin}. Despite the lack of detailed diversity metrics for each skin tone, the equal representation of all skin tones within the BalancedFace-4 offers us a reasonable basis to regard it as more representative of the global population compared to other datasets in the literature.

\subsection{Training Algorithm}

As described in Sec.~\ref{sec:problem_form}, we exploit the features extracted by a face recognition model for the face verification task. To this end, we opt for ArcFace \cite{deng2019arcface} as the training algorithm for the face recognition model. ArcFace is a widely applied approach, which is considered state-of-the-art for face recognition tasks. It utilizes an additive angular margin loss function that encourages intra-class compactness and inter-class discrepancy, which enhances its ability to generalize well to unseen identities. In particular, the additive angular margin loss can be defined as follows:
\begin{equation}
L = -\frac{1}{N}\sum^{N}_{i=1}\log\frac{e^{s(\cos(\theta_{y_i} + m))}}{e^{s(\cos(\theta_{y_i} + m))} + \sum_{j=1, j \neq y_i}^{N}e^{s\cos(\theta_{j})}},
\end{equation}
where $s$ is a scaling factor, $m$ is the additive margin parameter, $\theta_{y_i}$ is the angle between the ground truth class center and the feature $\mathbf{X}_i$, and $N$ is the total number of images.
By leveraging the ArcFace algorithm, we anticipate training an effective baseline model that will ensure the soundness of the following fairness analysis.

\subsection{Evaluation Benchmark}

We evaluate on the RFW \cite{wang2019racial} benchmark due to its broad assortment of images in terms of race. Nevertheless, RFW only provides racial annotation, while age and gender annotations are necessary for the objectives of this work. To this end, the Amazon Rekognition\footnote{\url{https://aws.amazon.com/rekognition/}} service  was used to annotate the samples with gender and age labels.
To ensure the  reliability of this analysis, a manual validation step was performed in order to evaluate Amazon Rekognition's performance in extracting gender and age labels. For this purpose, we randomly sampled 50 images per gender and age group. Each image was manually annotated w.r.t. gender and age by one annotator, and the resulting labels were compared with those predicted by the Rekognition service. Gender attribute consists of two classes (male and female), while 6 age groups represent the age classes, namely 0-20, 21-30, 31-40, 41-50, 51-60, and 61-100.
The results of this validation process were organized in Tables~\ref{tab:rek_val_gender} and \ref{tab:rek_val_age}, presenting a clear overview of the model's annotation performance w.r.t. gender and age, respectively. 
\begin{table}[t]

  \begin{minipage}[t]{0.5\linewidth}
    \centering
    \caption{Performance of Amazon Rekognition API on providing gender annotations for RFW benchmark.}
    % \resizebox{\linewidth}{!}{
    % Second table
    \begin{tabular}{|c|c|c|c|c|}
    \hline
    \multirow{2}{*}{Age} & \multicolumn{4}{c|}{FPR} \\
    \cline{2-5}
    & Caucasian & Asian & Indian & African \\
    \hline
    Female & 0.00 & 0.02 & 0.04 & 0.07 \\
    Male & 0.00 & 0.00 & 0.00 & 0.00 \\
    \hline
    \end{tabular}%}
    \label{tab:rek_val_gender}
  \end{minipage}
  % \hfill
  \begin{minipage}[t]{0.5\linewidth}
  \caption{Performance of Amazon Rekognition API on providing age annotations for RFW benchmark.}
    \centering
    % Third table
    % \resizebox{\linewidth}{!}{
    \begin{tabular}{|c|c|c|c|c|}
    \hline
    \multirow{2}{*}{Age} & \multicolumn{4}{c|}{FPR} \\
    \cline{2-5}
    & Caucasian & Asian & Indian & African \\
    \hline
    0-20 & 0.06 & 0.02 & 0.1 & 0.06 \\
    21-30 & 0.02 & 0.04 & 0.04 & 0.04 \\
    31-40 & 0.02 & 0.12 & 0.14 & 0.02 \\
    41-50 & 0.06 & 0.06 & 0.12 & 0.08 \\
    51-60 & 0.06 & 0.1 & 0.04 & 0.1 \\
    61-100 & 0.04 & 0.0 & 0.04 & 0.08 \\
    \hline
    \end{tabular} %}
    
    \label{tab:rek_val_age}
  \end{minipage}
\end{table}

As reported in Tab.~\ref{tab:rek_val_gender}, there are only subtle differences in FPR values between different genders and races. For instance, among the female individuals, the FPR is highest for the African race (0.07), followed by Indian (0.04), Asian (0.02), and Caucasian (0.0). However, for male individuals, the FPR is consistently 0.0 across all racial categories. This indicates that Amazon Rekognition may exhibit small gender-related biases, particularly for female individuals of certain racial backgrounds. However, the overall quality of the annotations is high enough for the purpose of our analysis.

In Tab.~\ref{tab:rek_val_age}, we can also observe discrepancies across age groups and races. For example, in the age range of 0-20, the FPR is highest for the Indian race (0.1), followed by Caucasian (0.06), African (0.06), and Asian (0.02). In the age range of 31-40, the FPR is highest for the Indian race (0.14), followed by Asian (0.12), African (0.02), and Caucasian (0.02). However, it is noteworthy that almost all the false positives belong to neighboring age groups. Thus, we can infer that these errors do not severely affect our fairness assessment w.r.t. age.

% \begin{table}[h]
% \centering
% \caption{Performance of Amazon Rekognition API on providing gender annotations for RFW benchmark.}

% \begin{tabular}{|c|c|c|c|c|}
% \hline
% \multirow{2}{*}{Age} & \multicolumn{4}{c|}{FPR} \\
% \cline{2-5}
%  & Caucasian & Asian & Indian & African \\
% \hline
% Female & 0.00 & 0.02 & 0.04 & 0.07 \\
% Male & 0.00 & 0.00 & 0.00 & 0.00 \\
% \hline
% \end{tabular}
% \label{tab:rek_val_gender}
% \end{table}

% \begin{table}[h]
% \centering
% \caption{Performance of Amazon Rekognition API on providing age annotations for RFW benchmark.}

% \begin{tabular}{|c|c|c|c|c|}
% \hline
% \multirow{2}{*}{Age} & \multicolumn{4}{c|}{FPR} \\
% \cline{2-5}

%  & Caucasian & Asian & Indian & African \\
% \hline
% 0-20 & 0.06 & 0.02 & 0.1 & 0.06 \\
% 21-30 & 0.02 & 0.04 & 0.04 & 0.04 \\
% 31-40 & 0.02 & 0.12 & 0.14 & 0.02 \\
% 41-50 & 0.06 & 0.06 & 0.12 & 0.08 \\
% 51-60 & 0.06 & 0.1 & 0.04 & 0.1 \\
% 61-100 & 0.04 & 0.0 & 0.04 & 0.08 \\
% \hline
% \end{tabular}
% \label{tab:rek_val_age}
% \end{table}

\subsection{Evaluation metrics}
The typical metrics involved in the conducted analysis are accuracy, TPR, FPR, and mAP.
% \begin{equation}
% accuracy = P(\hat{y_i} = y_i),
% \end{equation}
% \begin{equation}
% TPR = P(\hat{y_i} = 1 \mid y_i = 1),
% \end{equation}
% \begin{equation}
% FPR = P(\hat{y_i} = 1 \mid y_i = 0),
% \end{equation}
% \begin{equation}
% mAP = \frac{1}{N} \sum_{q=1}^{N} \left( \frac{1}{N-1} \sum_{k=1}^{N-1} Precision(R_{kq}) \right),
% \end{equation}
% where $N$ is the number of queries (or images), $N-1$ is the number of relevant images for query $q$, and $Precision(R_{kq})$ is the proportion of relevant images among the top $k$ images retrieved for query $q$.
Moreover, we used the sum of absolute values of DFPR and DFNR to measure the total disparate mistreatment \cite{krasanakis2018adaptive}:
\begin{equation}
    D_M = \mid D_{FPR} \mid + \mid D_{FNR} \mid,
\end{equation}
where
\begin{equation}
D_{FPR}  = P\left(y_i=0 \mid \hat{y}_i=1, t_i=1 \right) -P\left(y_i=0 \mid \hat{y}_i=1, t_i=0 \right), 
\end{equation}
\begin{equation}
   D_{FNR}  = P\left(y_i=1 \mid \hat{y}_i=0, t_i=1 \right)  -P\left(y_i=1 \mid \hat{y}_i=0, t_i=0 \right),
\end{equation}
where $\hat{y_i}$ denotes the predicted label of sample $i$ and $t_i$ denotes whether sample $i$ belongs to a certain protected group (or sub-group) or not. 
For measuring the disparate impact, we use the p\%-rule, a fairness metric that ensures both the positive classification rates among any two groups are similar. Specifically, the smaller rate should be at least p\% of the larger one:
\begin{equation}
    \min \left \{ \frac{P(\hat{y_i}=1\mid t_i=0)}{P(\hat{y_i}=1\mid t_i=1)}, \frac{P(\hat{y_i}=1\mid t_i=1)}{P(\hat{y_i}=1\mid t_i=0)} \right \} \geq \frac{p}{100}.
\end{equation}
In a fair model as per p\%-rule, the above formula holds for p equal to 100. However, in practice, the threshold for $p$ may vary. For instance, the US Equal Opportunity Commission suggests a threshold of 80\% \cite{biddle2006adverse}.

\subsection{Implementation details}
In terms of pre-processing, we employ an MTCNN model \cite{zhang2016mtcnn} to detect and crop the faces and subsequently resize them to a dimension of 112$\times$112. To normalize the pixel intensities, we subtract a value of 127.5 from each pixel and subsequently divide the result by 128.
Regarding the training procedure, we employ the SGD optimizer with momentum 0.9, weight decay 5e-4, and an initial learning rate of 0.001 that decays by a factor of 10 at 70,000, 100,000, and 120,000 iterations, while the total training iterations are 145,000. Regarding the network architecture, we opted for training from scratch a ResNet-34, following the setup of other works \cite{wang2021meta,wang2020mitigating,wang2019racial}. The batch size is set to 256. For evaluation, the official test subsets (one for each race) and protocol of RFW \cite{wang2019racial} were used. Experiments were conducted on a single NVIDIA RTX-3090 GPU.

% split results into subsections
%map 
%prule, dfnr, dfr
%visualization / inter intra
\section{Results}

%intersections classic metrics acc, tpr fpr
\subsection{Evaluation using Typical Face Verification Metrics}
To begin with, it is important to consider the overall performance of the system as a function of race, as indicated in Tab.~\ref{tab:results_race_gender}. The model's accuracy ranges between 92.40\% and 95.91\%, and the TPR@FPR=1\% metric falls between 80.1\% and 91.35\%. The model performs best for the Caucasians, with accuracy and TPR@FPR=1\% of 95.91\% and 91.35\% respectively, and worst for the Africans, with values of 92.40\% and 80.10\%. Here, we observe that although performance discrepancies in terms of accuracy are low, the corresponding TPR disparities are much higher revealing the biased behavior of the model w.r.t. race. 
% \begin{table}[t]
% \centering
% \caption{Overall performance on RFW.}
% \label{tab:results_overall}
% \begin{tabular}{|c|c|l|}
% \hline
% Race & Accuracy & TPR@FPR=1\% \\
% \hline
% Caucasian & 0.9591 $\pm$ 0.0069 & 0.9135 \\
% African & 0.9240 $\pm$ 0.0081 & 0.8010 (-12.3\%)\\
% Asian & 0.9301 $\pm$ 0.0089 & 0.8206 (-10.2\%)\\
% Indian & 0.9441 $\pm$ 0.0104 & 0.8847 (-3.15\%)\\
% \hline
% \end{tabular}
% \end{table}

\begin{table}[t]
\centering
\caption{Overall and Race-Gender intersection performance on RFW.}
\label{tab:results_race_gender}
% \resizebox{1\linewidth}{!}{
\begin{tabular}{|c|c|l|c|c|l|}
\hline
Race & Accuracy & TPR@FPR=1\% & Gender & Accuracy & TPR@FPR=1\% \\
\hline
\multirow{2}{*}{Caucasian} & \multirow{2}{*}{0.9591 $\pm$ 0.0069} & \multirow{2}{*}{0.9135} & Male & 0.9636 $\pm$ 0.0031 & 0.9152 \\
&&& Female & 0.9552 $\pm$ 0.0069 & 0.9065 (-0.9\%)\\
\hline
\multirow{2}{*}{African} & \multirow{2}{*}{0.9240 $\pm$ 0.0081} & \multirow{2}{*}{0.8010 (-12.3\%)} & Male & 0.9231 $\pm$ 0.0076 & 0.8560  \\
&&& Female & 0.9193 $\pm$ 0.0268 & 0.8091 (-5.4\%) \\
\hline
\multirow{2}{*}{Asian} & \multirow{2}{*}{0.9301 $\pm$ 0.0089} & \multirow{2}{*}{0.8206 (-10.2\%)} & Male & 0.9420 $\pm$ 0.0036 & 0.8621   \\
&&& Female & 0.9043 $\pm$ 0.0043 & 0.7449 (-13.6\%) \\
\hline
\multirow{2}{*}{Indian} & \multirow{2}{*}{0.9441 $\pm$ 0.0104} & \multirow{2}{*}{0.8847 (-3.15\%)} & Male & 0.9524 $\pm$ 0.0060 & 0.8983  \\
&&& Female & 0.9304 $\pm$ 0.0165 & 0.8581 (-4.5\%) \\
\hline
\end{tabular}%}
\end{table}
Taking a more nuanced look at the performance by considering the intersection of race and gender in Tab.~\ref{tab:results_race_gender}, we see varying results. Within each ethnic group, there are noticeable differences between male and female accuracy and TPR@FPR=1\% scores. Notably, in all the races, males outperform females in both accuracy and TPR@FPR=1\%, while the discrepancies w.r.t. TPR are again much larger. For instance, the TPR@FPR=1\% for Asian Males is 11,72\% higher compared to Asian Females. 
Similarly, Tab.~\ref{tab:results_race_age} presents the model's performance on race-age intersections. The Caucasian and Indian groups demonstrate high accuracy and TPR@FPR=1\% across the age groups of 41-50 and 61-60, while TPR@FPR=1\% for Indians aged between 0-20 is considerably lower, i.e., 0.79. The African and Asian groups exhibit more variability across age ranges, most notably the African and Asian groups experience a sharp decrease in both accuracy and TPR@FPR=1\% in the oldest and the youngest age intervals, respectively. These findings indicate the importance of assessing the model's performance on multiple sensitive attributes when evaluating it in terms of fairness.

\begin{table}[t]
\centering
\caption{Performance at the intersection of Race and Age on RFW.}
\label{tab:results_race_age}
% \resizebox{0.65\linewidth}{!}{
\begin{tabular}{|c|c|c|l|}

\hline
\textbf{Race} & Age Range & Acc $\pm$ Std & TPR@FPR=1\% \\
\hline
\multirow{6}{*}{Caucasian} & 0-20 & 0.9366 $\pm$ 0.0323 & 0.9076 (-3.7\%)\\
& 21-30 & 0.9569 $\pm$ 0.0086 & 0.9057 (-3.9\%)\\
& 31-40 & 0.9550 $\pm$ 0.0010 & 0.9136 (-3.1\%) \\
& 41-50 & 0.9641 $\pm$ 0.0090 & 0.9430 \\
& 51-60 & 0.9643 $\pm$ 0.0083 & 0.9416 (-0.1\%)\\
& 61-100 & 0.9562 $\pm$ 0.0135 & 0.9134 (-3.1\%)\\
\hline
\multirow{6}{*}{African} & 0-20 & 0.9123 $\pm$ 0.1241 & 0.9167 \\
& 21-30 & 0.9261 $\pm$ 0.0099 & 0.8029 (-12.4\%) \\
& 31-40 & 0.9301 $\pm$ 0.0038 & 0.8225 (-10.3\%)\\
& 41-50 & 0.9276 $\pm$ 0.0098 & 0.8275 (-9.7\%)\\
& 51-60 & 0.9506 $\pm$ 0.0243 & 0.9084 (-0.9\%)\\
& 61-100 & 0.8306 $\pm$ 0.0571 & 0.7980 (-12.9\%)\\
\hline
\multirow{6}{*}{Asian} & 0-20 & 0.9110 $\pm$ 0.0203 & 0.7625 (-16.5\%)\\
& 21-30 & 0.9207 $\pm$ 0.0140 & 0.8050 (-11.9\%)\\
& 31-40 & 0.9308 $\pm$ 0.0156 & 0.8498 (-7.0\%)\\
& 41-50 & 0.9651 $\pm$ 0.0099 & 0.8980 (-1.7\%)\\
& 51-60 & 0.9528 $\pm$ 0.0069 & 0.8763 (-4.1\%)\\
& 61-100 & 0.9415 $\pm$ 0.0214 & 0.9135 \\
\hline
\multirow{6}{*}{Indian} & 0-20 & 0.9222 $\pm$ 0.0222 & 0.7903 (-14.6\%)\\
& 21-30 & 0.9313 $\pm$ 0.0054 & 0.8684 (-6.2\%)\\
& 31-40 & 0.9434 $\pm$ 0.0193 & 0.8586 (-7.2\%)\\
& 41-50 & 0.9583 $\pm$ 0.0038 & 0.9255 \\
& 51-60 & 0.9575 $\pm$ 0.0013 & 0.9242 (-0.1\%)\\
& 61-100 & 0.9247 $\pm$ 0.0317 & 0.8318 (-10.1\%)\\
\hline
\end{tabular}%}
\end{table}

\begin{table}[h]
\caption{Cross-age performance on RFW.}
\centering
%\resizebox{0.6\linewidth}{!}{
\begin{tabular}{|c|c|c|l|}
\hline
Race & Age Gap & Acc $\pm$ Std & TPR@FPR=1\% \\
\hline
\multirow{5}{*}{Caucasian} & 0 & 0.9663 $\pm$ 0.0048 & 0.9397 \\
& 1 & 0.9580 $\pm$ 0.0043 & 0.9142 (-2.7\%) \\
& 2 & 0.9511 $\pm$ 0.0036 & 0.8767 (-6.7\%) \\
& 3 & 0.9570 $\pm$ 0.0054 & 0.7792 (-17.1\%) \\
& 4 & 0.9141 $\pm$ 0.0234 & 0.7222 (-23.1\%) \\
\hline
\multirow{5}{*}{African} & 0 & 0.9272 $\pm$ 0.0052 & 0.8132 (-1.3\%)\\
& 1 & 0.9254 $\pm$ 0.0004 & 0.8239 \\
& 2 & 0.9343 $\pm$ 0.0081 & 0.7041 (-14.5\%)\\
& 3 & 0.9083 $\pm$ 0.0583 & 0.7440 (-9.7\%)\\
& 4 & 0.7500 $\pm$ 0.0833 & 0.0000 (-100\%)\\
\hline
\multirow{5}{*}{Asian} & 0 & 0.9235 $\pm$ 0.0020 & 0.7928 (-12.1\%) \\
& 1 & 0.9326 $\pm$ 0.0057 & 0.8445 (-6.3\%) \\
& 2 & 0.9290 $\pm$ 0.0040 & 0.7215 (-20.0\%) \\
& 3 & 0.9690 $\pm$ 0.0173 & 0.9018 \\
& 4 & 0.8966 $\pm$ 0.0000 & 0.4722 (-47.6\%) \\
\hline
\multirow{5}{*}{Indian} & 0 & 0.9512 $\pm$ 0.0011 & 0.9192 \\
& 1 & 0.9463 $\pm$ 0.0026 & 0.8834 (-3.9\%) \\
& 2 & 0.9355 $\pm$ 0.0155 & 0.7963 (-13.4\%) \\
& 3 & 0.8723 $\pm$ 0.0215 & 0.6061 (-34.1\%) \\
& 4 & 0.8181 $\pm$ 0.0033 & 0.5222 (-43.2\%) \\
\hline
\end{tabular}%}
\label{tab:age_gap}
\end{table}

\begin{table}[t]

  \begin{minipage}[t]{0.6\linewidth}
    \centering
    \caption{Retrieval performance across demographics on the RFW benchmark. The reported TPR values correspond to an FPR of 0.5\%.}
    %\resizebox{\linewidth}{!}{
    % Second table
    \begin{tabular}{|c|c|l|l|}
\hline
Attribute & Group & mAP & TPR  \\
\hline
\multirow{4}{*}{Race} & Caucasian & 0.7694 (-1.7\%) & 0.9634 \\
 & Asian &   0.7253 (-7.3\%) & 0.9527  (-1.1\%) \\
 & Indian &  0.7825  &         0.9509  (-1.3\%) \\
 & African & 0.7455 (-4.7\%) & 0.9551  (-0.9\%) \\
\hline
\multirow{2}{*}{Gender}  & Male & 0.7672 & 0.9556 \\
& Female & 0.7247 (-5.5\%) & 0.9554 (-0.0\%) \\

\hline
\multirow{6}{*}{Age} & 0-20 & 0.6516 (-19.1\%) & 0.9431 (-1.9\%) \\
 & 21-30 & 0.7366 (-8.5\%) & 0.9550 (-0.7\%)  \\
 & 31-40 & 0.7614 (-5.4\%)  & 0.9545 (-0.8\%) \\
 & 41-50 & 0.7797 (-3.2\%) & 0.9572 (-0.5\%) \\
 & 51-60 & 0.8053 & 0.9619 \\
 & 61-100 & 0.7924 (-1.6\%) & 0.9566  (-0.5\%) \\
\hline
\end{tabular}%}
    \label{tab:map}
  \end{minipage}
  % \hfill
  \begin{minipage}[t]{0.4\linewidth}
  \caption{Inter- and Intra-group cosine similarities with respect to protected groups.}
    \centering
    % Third table
    \resizebox{\linewidth}{!}{
    \begin{tabular}{|c|c|c|}
\hline
\multirow{2}{*}{Group} & \multicolumn{2}{c|}{Mean ($\times10^{-3}$) $\pm$ Std ($\times10^{-4}$)} \\ \cline{2-3}
& Inter-group & Intra-group \\ \hline
\multicolumn{3}{|c|}{Races} \\ \hline
Caucasian & 0.97 $\pm$ 68.33 & 1.76 $\pm$ 78.93 \\
Asian & 3.00 $\pm$ 67.92 & 8.81 $\pm$ 79.05 \\
Indian & 3.04 $\pm$ 68.89 & 31.62 $\pm$ 77.92 \\
African & -0.15 $\pm$ 67.44 & 19.98 $\pm$ 90.85 \\ \hline
\multicolumn{3}{|c|}{Genders} \\ \hline
Female & 3.92 $\pm$ 67.83 & 7.27 $\pm$ 79.72 \\
Male & 3.92 $\pm$ 67.83 & 5.92 $\pm$ 74.45 \\ \hline
\multicolumn{3}{|c|}{Age} \\ \hline
0-20 & 4.12 $\pm$ 70.11 & 15.18 $\pm$ 84.25 \\
21-30 & 4.43 $\pm$ 71.05 & 7.74 $\pm$ 75.83 \\
31-40 & 4.40 $\pm$ 72.18 & 6.69 $\pm$ 74.79 \\
41-50 & 4.68 $\pm$ 71.47 & 9.16 $\pm$ 75.49 \\
51-60 & 4.41 $\pm$ 70.36 & 9.61 $\pm$ 78.31 \\
61-100 & 3.27 $\pm$ 70.06 & 17.96 $\pm$ 86.50 \\ \hline
\end{tabular} }
    
    \label{tab:inter_intra_group}
  \end{minipage}
\end{table}

Another important but often overlooked problem is the effectiveness of face verification models on image pairs with large age gaps. Tab.~\ref{tab:age_gap} reports the results obtained on this setting and uncovers the system's decreasing effectiveness when the age gap between the two images increases. Interestingly, the Caucasian group maintains relatively high performance across age gap categories (i.e., 0.2175 maximum differences in terms of TPR), whereas the African group experiences the most drastic drop in performance (i.e., -0.1843 in terms of accuracy and -0.8132 in terms of TPR). Overall, the performance decreases for larger Age Gap values for all the race groups except for Asians.  Note that Age Gap values in Tab.~\ref{tab:age_gap} denote the distance between the Age labels of the images within a pair.

%map
%Evaluating a face verification system in terms of fairness 
% Furthermore, it has value to evaluate the model's effectiveness in retrieving specific identities from the full dataset.
Furthermore, evaluating the model's capability to retrieve specific identities from the entire dataset is essential.
% Furthermore, in 
Tab.~\ref{tab:map} 
% we present 
presents the retrieval performance considering all the test samples as queries and documents. TPR does not reflect the model's bias due to the numerous easy negatives added in this experimental setup. However, mAP values fluctuate across different groups. Notably, the mAP for Males is 4.2\% higher than Females. In addition, the age group 51-60 is associated with the highest mAP, while the age group 0-20 with the lowest.

It is evident that the face verification model performs better on some racial groups than others, with Caucasians consistently having the best results across all metrics, except for mAP. 
Given that the training data is balanced w.r.t. race, these disparities can be attributed to the lack of diverse representations in the training dataset or to inherent visual characteristics of particular groups typically ignored during training \cite{georgopoulos2020investigating,cavazos2020accuracy}.
Similarly, gender-based performance variations are conspicuous. Men are more accurately recognized than women in all racial groups. The under-representation of females in the training data can be considered the prime reason for the model's discriminatory behavior \cite{hupont2019demogpairs}.
Analyzing the effect of age brings more depth to our understanding of the system’s performance. In general, there is no consistent performance trend to let us interpret the model's behavior, probably due to the high imbalance of RFW w.r.t. ages representation \cite{wang2019racial}. However, TPR is considerably low for the oldest in the African and Indian groups, which could be attributed to the more pronounced age-related morphological changes, making older individuals harder to recognize. The age group 0-20 is also associated with lower performance for Asians and Indians, potentially due to the rapid physical changes experienced during these years, and the relative under-representation of this age group in the training data.

\subsection{Feature Space Exploration}
Here, we explore the model's feature space through t-SNE visualizations and the inter- and intra-group cosine similarities between different protected groups to gain a better understanding of the model's behavior. 
Tab.~\ref{tab:inter_intra_group} presents the inter- and intra-group cosine similarities concerning races, genders, and ages.
Regarding race, it is interesting to note that Caucasians and Africans exhibit lower inter-group cosine similarity, indicating that these groups are well-separated in the feature space, potentially because their skin tones provide clear distinctions compared to other races. Although these observations do not constitute a sufficient basis for bias on their own, when we delve into intra-group similarities, a different picture emerges. 
The African group has a higher intra-group similarity compared to the Caucasian group, suggesting the model struggles to differentiate between African individuals, resulting in the higher FPR observed for this demographic. 
Such findings are also noticed for Asians and Indians. Similar patterns are observed for the different gender and age groups.In particular, women exhibit higher intra-group similarities, while  higher intra-group cosine similarities emerge predominantly at the lower and upper ends of the age spectrum.

\begin{figure}[h]
\resizebox{1\linewidth}{!}{
\begin{tabular}{ccc}
\centering
  \includegraphics[width=0.33\linewidth, trim=2cm 1cm 2cm 1cm, clip]{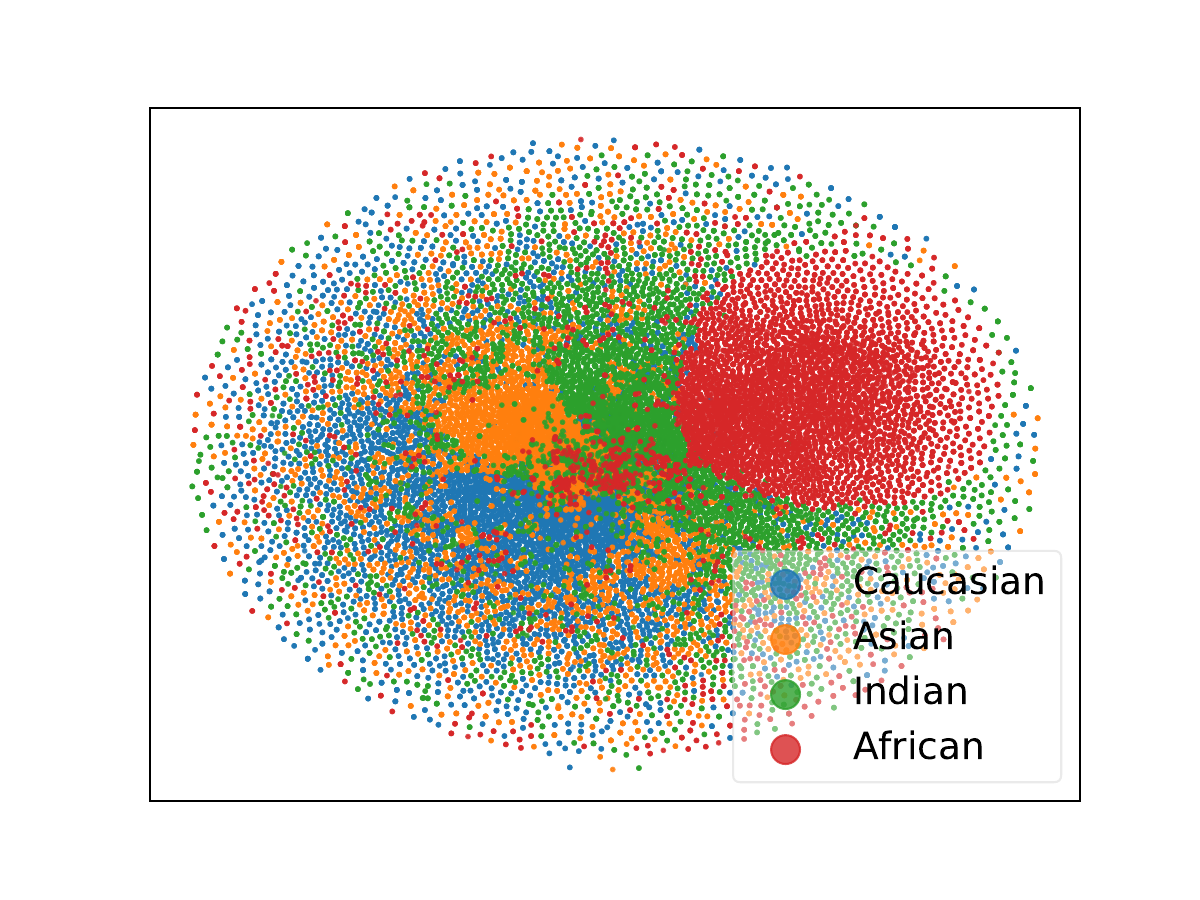} &   \includegraphics[width=0.33\linewidth, trim=2cm 1cm 2cm 1cm, clip]{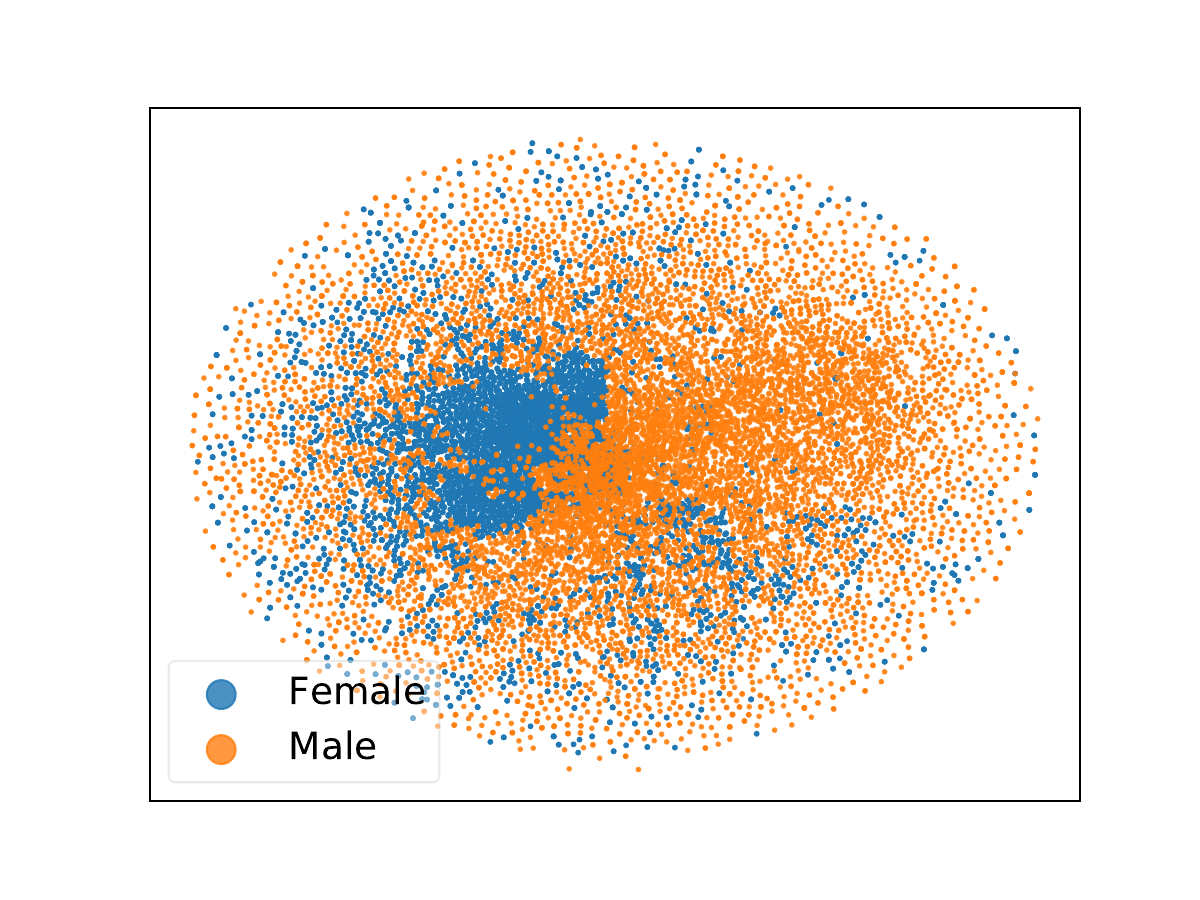} &  \includegraphics[width=0.33\linewidth, trim=2cm 1cm 2cm 1cm, clip]{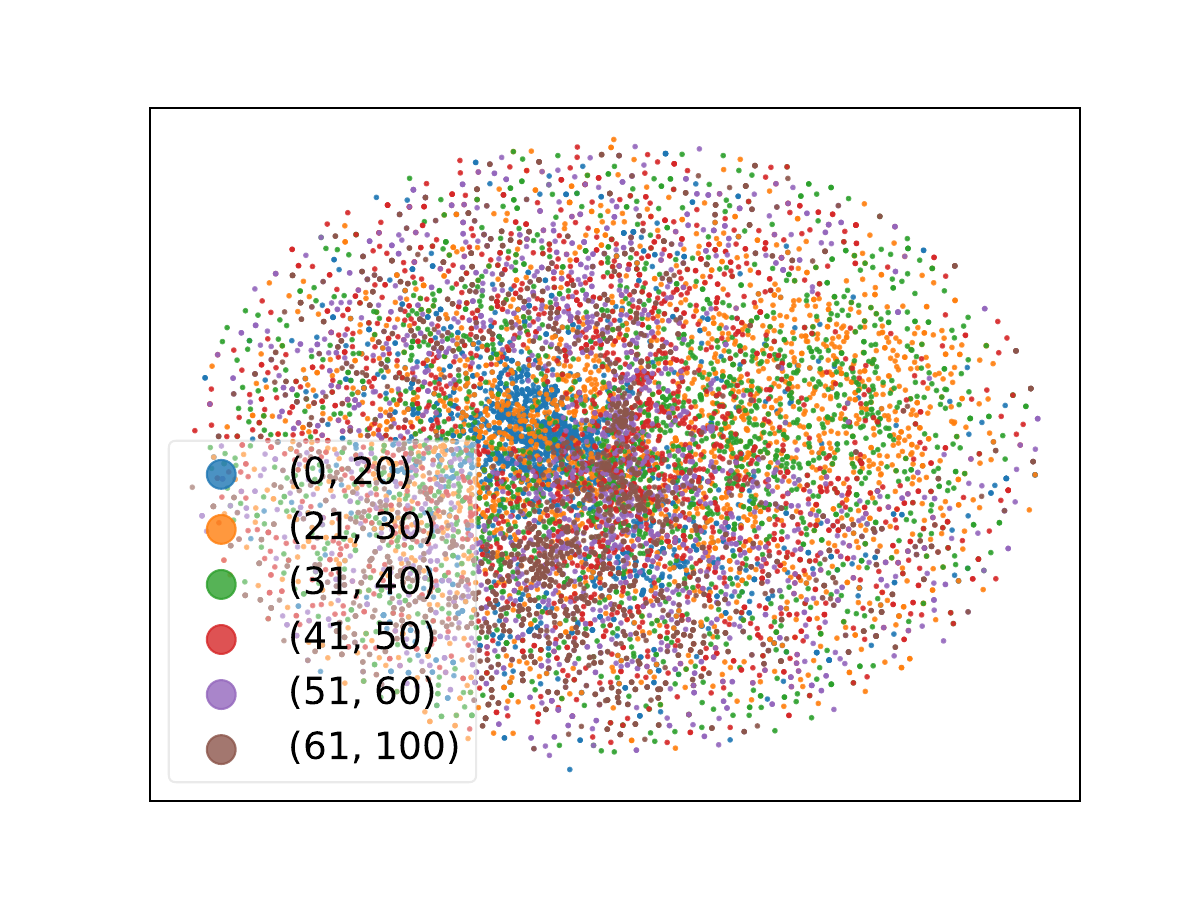}\\
(a) Race & (b) Gender & (c) Age \\
\end{tabular}}
\caption{t-SNE visualizations w.r.t. demographics.}
\label{fig:tsne}
\end{figure}

% tsne inter intra
These findings are also reflected by the t-SNE visualizations of Fig.~\ref{fig:tsne} for race, gender, and age groups. Particularly, in Fig.~\ref{fig:tsne}(a), Caucasians are positioned on the left and Africans on the right, while the African, Indian, and Asian clusters appear more compact compared to the Caucasian cluster. In Fig.~\ref{fig:tsne}(b), the cluster of females is significantly more compact compared to the one of males. 
Finally, in Fig.~\ref{fig:tsne}(c) we observe that groups at the lower and higher ends of the age spectrum exhibit more compact distributions and reside in close proximity. 

  \begin{table}[t]
  \caption{Race-Gender-Age intersections}
    \centering
    % Third table
    \resizebox{1\linewidth}{!}{
    \begin{tabular}{|c|c|c|l|c?c|c|c|l|c|}
        \hline
        Race & Gender & Age & p\%-rule & $D_M$ &  Race & Gender & Age & p\%-rule & $D_M$ \\
        \hline
        \multirow{12}{*}{Caucasian} & \multirow{6}{*}{Male} & 0-20 & 0.8347 (-16.0\%) & 0.0505  &  \multirow{12}{*}{Asian} & \multirow{6}{*}{Male} & 0-20 & 0.9098 (-8.5\%) & 0.0447\\
         &  & 21-30 & 0.9741 (-1.9\%) & 0.0412  & &  & 21-30 & 0.9568 (-3.8\%) & \textbf{0.0502} \\
         &  & 31-40 & 0.9711 (-2.2\%) & 0.0487  & &  & 31-40 & 0.9947 & 0.0453 \\
         &  & 41-50 & 0.9796 (-1.4\%) & \textbf{0.0772} & &  & 41-50 & 0.9136 (-8.1\%) & 0.0453 \\
         &  & 51-60 & 0.9671 (-2.6\%) & 0.0684  & &  & 51-60 & 0.9927 (-0.2\%) & 0.0345 \\
         &  & 61-100 & 0.9785 (-1.5\%) & 0.0385  & &  & 61-100 & 0.9721 (-2.3\%) & 0.0159 \\ \cline{2-5} \cline{7-10}
         & \multirow{6}{*}{Female} & 0-20 & 0.9478 (-4.6\%) & 0.0351  & & \multirow{6}{*}{Female} & 0-20 & 0.8478 (-14.8\%) & \textbf{0.1252} \\
         &  & 21-30 & 0.9341 (-5.9\%) & 0.0583  & &  & 21-30 & 0.9296 (-6.5\%) & 0.0631 \\
         &  & 31-40 & 0.9932 & 0.0271  & &  & 31-40 & 0.9149 (-8.0\%) & 0.0551 \\
         &  & 41-50 & 0.8843 (-11.0\%) & \textbf{0.0639} & &  & 41-50 & 0.9592 (-3.6\%) & 0.0849 \\
         &  & 51-60 & 0.9746 (-1.9\%) & 0.0169 & &  & 51-60 & 0.9908 (-0.4\%) & 0.0743 \\
         &  & 61-100 & 0.9259 (-6.8\%) & 0.0630  & &  & 61-100 & 0.9407 (-5.4\%) & 0.1237 \\
\hline \hline
        
        \multirow{12}{*}{Indian} & \multirow{6}{*}{Male} & 0-20 & 0.9659 (-2.9\%) & \textbf{0.0802} & \multirow{12}{*}{African} & \multirow{6}{*}{Male} & 0-20 & 0.8612 (-13.1\%) & 0.0938 \\
         &  & 21-30 & 0.9500 (-4.5\%) & 0.0368 & &  & 21-30 & 0.8902 (-10.2\%) & \textbf{0.1043} \\
         &  & 31-40 & 0.9651 (-3.0\%) & 0.0361 & &  & 31-40 & 0.9414 (-5.0\%) & 0.0478 \\
         &  & 41-50 & 0.9566 (-3.8\%) & 0.0218 & &  & 41-50 & 0.9632 (-2.8\%) & 0.0203 \\
         &  & 51-60 & 0.9949  & 0.0580 & &  & 51-60 & 0.9404 (-5.1\%) & 0.0358 \\
         &  & 61-100 & 0.9778 (-1.7\%) & 0.0076 & &  & 61-100 & 0.9353 (-5.6\%) & 0.1017 \\ \cline{2-5} \cline{7-10}
         & \multirow{6}{*}{Female} & 0-20 & 0.9765 (-1.8\%) & 0.0432 & & \multirow{6}{*}{Female} & 0-20 & 0.9719 (-1.9\%) & 0.0872 \\
         &  & 21-30 & 0.9870 (-0.8\%) & 0.0320 & &  & 21-30 & 0.9285 (-6.3\%) & 0.0495 \\
         &  & 31-40 & 0.9439 (-5.1\%) & 0.0072 & &  & 31-40 & 0.9909 & 0.0677 \\
         &  & 41-50 & 0.9717 (-2.3\%) & 0.0325 & &  & 41-50 & \underline{0.7836} (-20.9\%) & 0.0916 \\
         &  & 51-60 & 0.9486 (-4.6\%) & 0.0081 & &  & 51-60 & \underline{0.7482} (-24.5\%) & 0.1236 \\
         &  & 61-100 & \underline{0.6384} (-35.9\%) & \textbf{0.1262} & &  & 61-100 & \underline{0.6859} (-30.8\%) & \textbf{0.4761} \\
\hline

    \end{tabular} }
    
    \label{tab:3intersec}
  \end{table}

% prule dfnr dfpr
\subsection{Evaluation Using Fairness Metrics}
In this subsection we present the models' performance w.r.t. disparate impact and disparate mistreatment metrics. Examining the intersection of age, gender, and racial disparities reveals common patterns across racial groups, as presented in Tab.~\ref{tab:3intersec}. Among Caucasians, it is observed that the least disparity is evident in the segment of females aged 31-40, while males within the younger age range (i.e., 0-20) experience the most disparity. The Asian demographic paints a slightly different picture, with the least disparity appearing in males aged 31-40, contrasted by the highest disparity found among the younger female group, aged 0-20.
When considering the Indian demographic, the age-gender group exhibiting the lowest disparity is males aged 51-60, while the group facing the highest disparity is the oldest one, with the most prominent discrepancies found in females aged 61-100.
Lastly, within the African demographic, the results indicate that females aged 31-40 benefit from the lowest disparity, while, as with the Indian demographic, the highest level of disparity is seen among older females (61-100).

Overall, we can infer that the model mostly discriminates against females aged either 0-20 or 61-100. Moreover, there are four subgroups where the p\%-rule is below the critical fairness threshold of 80\%, namely the Indian Females aged 61-100 and African Females aged 41-50, 51-60, and 61-100. These insights reaffirm our hypothesis that solely relying on typical metrics such as accuracy fails to fully capture biases among various groups. As it has been shown, even a relatively small discrepancy in accuracy can correspond to a significant variation in multiple fairness metrics. This indicates the importance of using a broad range of indicators when assessing the performance and fairness of such models.

\section{Conclusion}
The analysis conducted in this paper uncovers significant biases present in state-of-the-art facial verification technology, highlighting the necessity for more inclusive and equitable algorithms. By employing a comprehensive approach combining quantitative analysis and t-SNE visualizations, our analysis demonstrated that these systems have a tendency towards misidentifying certain demographic groups, i.e., non-Caucasians, females, and youngest/oldest people. Furthermore, the conducted analysis using several fairness metrics highlights the inadequacy of accuracy as a fairness metric in face verification tasks. As facial recognition and verification technology is increasingly deployed in various critical sectors, such biases could have profound disparate impacts on people's lives. However, one limitation of our study is the absence of a fine-grained protected attributes analysis, a condition attributed to the lack of datasets offering such information in the existing literature. For instance, the category "Asian" encompasses a diverse group with distinct visual characteristics. This might obscure underlying biases and potential inequalities.  It is thus crucial that researchers, developers, and policy makers prioritize addressing these issues to ensure that such technology operates fairly and equitably for all individuals, regardless of their race, age, gender, or even other protected attributes.

%% ANONYMIZE
\section*{Acknowledgements} This research was supported by the EU Horizon Europe project
MAMMOth (Grant Agreement 101070285).

%Bibliography
\bibliographystyle{unsrt}  
\small
\bibliography{references}  

\end{document}